\pdfoutput=1
\documentclass[11pt]{article}

\usepackage[final]{emnlp/emnlp2021}

\usepackage{times}
\usepackage{latexsym}

\usepackage[utf8]{inputenc} % allow utf-8 input
\usepackage[T1]{fontenc}    % use 8-bit T1 fonts
\usepackage{hyperref}       % hyperlinks
\usepackage{url}            % simple URL typesetting
\usepackage{amsmath}
\usepackage{amsfonts}       % blackboard math symbols
\usepackage{nicefrac}       % compact symbols for 1/2, etc.a
\usepackage{microtype}      % microtypography
\usepackage{graphicx}
\usepackage{subfigure}
\usepackage{booktabs} % for professional tables
\usepackage{makecell}
\usepackage[singlelinecheck=false]{caption}
\usepackage{makecell}
\usepackage{listings}
\usepackage{xcolor}

\newcommand{\todo}[1]{{}}

\newcommand{\fixedcomment}[1]{{}}

\newcommand{\victor}[1]{{}}
\newcommand{\srush}[1]{{}}
\newcommand{\ella}[1]{{}}
\newcommand{\teven}[1]{{}}
\newcommand{\uurl}[1]{{\href{#1}{#1}}}

\title{Block Pruning For Faster Transformers}

\author{François Lagunas,
  Ella Charlaix, 
  Victor Sanh,
  Alexander M. Rush   
  
  \\ 
  
 Hugging Face \\ \texttt{\{francois, ella, victor, sasha\}@huggingface.co}}

\begin{document}

\maketitle

\begin{abstract}

  Pre-training has improved
  model accuracy for both classification and generation tasks at
  the cost of introducing much larger and slower models.  Pruning methods have proven to be an effective way
  of reducing model size, whereas distillation methods are proven for speeding up inference.
  We introduce a block
  pruning approach targeting both small and fast models. Our approach extends structured methods by considering blocks of any size and integrates this structure into the movement pruning
  paradigm for fine-tuning. 
  We find that this approach learns to prune out full components of
  the underlying model, such as attention heads. Experiments consider
  classification and generation tasks, yielding among other results a pruned
  model that is a 2.4x faster, 74\% smaller BERT on SQuAD v1, with a 1\% drop
  on F1, competitive both with distilled models in speed and pruned models in size.

\end{abstract}

\section{Introduction}

Pre-trained transformer models are the standard for NLP tasks in
both classification and generation tasks~\cite{devlin2018bert,lewis2019bart}. The recent trend is for
models to continue to grow in size while yielding improved performance
on standard benchmarks~\cite{rosset2020turing}. This development highlights the need to reduce
the storage size and increase the efficiency of pre-trained models.

Pruning methods have shown to be extremely effective at reducing the
storage size of models fine-tuned for a specific task. Approaches such
as magnitude pruning~\cite{DBLP:journals/corr/HanPTD15}, L0 regularization~\cite{LouizosWK18}, lottery ticket
hypothesis~\cite{LotteryTicketHypothesis}, diff pruning~\cite{Guo2020DiffPruning}, and movement pruning~\cite{sanh2020movement}
have demonstrated remarkable reductions in model size. Movement pruning  produces 77\%
savings in parameter storage for a 1\% drop in accuracy on SQuAD v1.1. However, these
models yield very little actual efficiency benefits, as to run them in
standard hardware often requires reconstructing the original dense
shape. 

On the other hand distillation methods have been more effective at producing faster models as has been shown by DistilBERT \cite{sanh2019distilbert}, TinyBERT \cite{jiao2019tinybert} or MobileBERT \cite{sun2020mobilebert}. These approaches utilize targeted distillation to produce smaller models with a dense structure that is fast on standard hardware. However without careful engineering and size selection these models are much larger than pruned ones. 

In this work, we target closing this gap through block pruning. 
Unlike pruning individual parameters, this approach
encourages pruning that can be optimized on dense hardware.
It is a less rigid approach than row or column-based pruning typically used in structured approaches~\cite{StructuredPruningQA}, which have been difficult to apply effectively to transformers.
We integrate this approach with Movement pruning~\cite{sanh2020movement}, a simple
method for pruning pre-trained models during fine-tuning. The final
method\footnote{Available at \url{https://github.com/huggingface/nn\_pruning}} has few additional hyperparameters or training requirements.

% SRUSH -> Dropping this for now and integrating.
%Researchers have tried to bridge this gap using structured
%pruning for Transformers. Structured pruning methods focus on
%pruning grouped components of the model such as intermediate layers or
%full rows or columns. These approaches have the potential to produce
%small dense models, like with distillation, while still starting
%directly from the original pretrained model as in pruning. %Experimental
%results though are not competitive with other approaches.

Experiments consider a large variety of different
benchmark datasets comparing accuracy and efficiency. We find a
surprising result that despite utilizing sub-row square blocks during
training, the approach learns to eliminate full components of
the model, effectively dropping a large number of attention heads.
This effect allows the model to achieve speedups even beyond standard
structured pruning of feed-forward layers. 
Results show a 2.4x speedup on
SQuAD v1.1 with a 1\% drop of F1, and a 2.3x speedup on QQP with a 1\% loss of F1. Experiments on summarization also show a 1.39x speedup for an average of 2 points drop on all ROUGE metrics on CNN/DailyMail, and for a reduction of decoder weights of 3.5x.

% Block sparsity as opposed to unstructured sparsity makes a big difference in terms of. That’s because one of the most important factors for improving computation speed is data locality, especially with highly parallel hardware like GPUs. For example, linear algebra libraries compute matrix multiplication using large blocks (128*64 is a common size). The explanation is that at a micro level those machines are using typically 32 ways SIMD, and memory must be loaded by these large contiguous chunks not to become the bottleneck. So you have to use blocks large enough to keep all the units and memory bus busy. This explains why unstructured sparsity is hard to implement with dense algebra performance on GPUs. Data locality is important for CPUs too, but the memory access capabilities and the general structure limits the difference between dense and sparse performance.

\section{Related Work}

There has been a growing interest in the compression of pre-trained language models. We consider three varieties of methods: distillation, pruning, and structured pruning.
% Existing compression techniques can be divided into four categories: knowledge distillation ~\citet{DBLP:journals/corr/HintonVD15}, weight pruning ~\citep{lecun1990optimal, DBLP:journals/corr/HanPTD15}, quantization ~\citep{courbariaux2015binaryconnect, Zhou, Jacob, Gong} and matrix factorization~\citep{Sainath, Xu, Chen, Winata}.

Knowledge distillation, introduced by~\citet{DBLP:journals/corr/HintonVD15}, is a popular compression technique. Researchers have applied this method 
to a variety of NLP models ~\citep{tang2019distilling, sun2019patient, turc2019well}.
Distillation has been used to obtain significantly smaller BERT models achieving competitive performances. \citet{sanh2019distilbert} distills BERT into shallower students during the pre-training stage and optionally during the fine-tuning stage. MobileBERT \citep{sun2020mobilebert} and TinyBERT \citep{jiao2019tinybert} are obtained thanks to a layer-wise distillation strategy. While the distillation of former is task-agnostic, the one used to obtain the latter is task-specific.

Other previous work has focused on unstructured pruning~\citep{lecun1990optimal, DBLP:journals/corr/HanPTD15,LotteryTicketHypothesis}. When targeting transformer models, it is typical to select the weights to prune based on their magnitude~\citep{gordon2020compressing}, or by computing an importance score using a first-order method~\citep{sanh2020movement}. While these methods allow for a significant reduction in model size, specialized hardware is required to make use of the resulting unstructured sparse matrices in order to speed up inference. 
 
In contrast, structured pruning removes coherent groups of weights ~\citep{murray2015auto, see2016compression, joulin2016fasttext, fan2019reducing, sajjad2020poor}. 
Recent works \citep{DBLP:journals/corr/abs-1905-10650, voita2019analyzing} show that some heads can be removed without significant degradation in performance, leading to the conclusion that most heads provide redundant information.  Other authors have worked on combining matrix factorization and weight pruning. While \citet{mao2020ladabert} combine SVD-based matrix factorization with unstructured pruning, \citet{wang2019structured} use structured pruning in order to reduce the rank. Related to our approach, \citet{kim2020fastformers} and \citet{StructuredPruningQA}  both apply structured pruning on the heads of the multi-head attention (MHA) and on the inner-layer nodes of the feed-forward network (FFN). The former uses predefined pruning ratios, shared across all layers, in order to select the modules to prune after sorting them given an importance score. \citet{StructuredPruningQA} compares different methods to compute the prunable module masks and find L0 regularization to perform the best.

\section{Background}

Starting with a
transformer model with parameters $\theta$,  our goal is to produce a
set of parameters $\theta'$ that are both fine-tuned for a specific
end-task and smaller in such a way that inference can be efficiently computed on parallel hardware.

The two largest  lines in the transformer parameter budget are the feed-forward network sub-layer (FFN) and the multi-head attention sub-layer (MHA). The FFN parameters consist of two matrices ($\mathbf{W_1}$ and $\mathbf{W_2}$) of transposed shape $\mathbb{R}^{d_{\mathrm{model}}\times d_{\mathrm{ff}}}$ and $\mathbb{R}^{d_{\mathrm{ff}}\times d_{\mathrm{model}}}$ where $d_{\mathrm{model}}$ is the hidden size and $d_{\mathrm{ff}} \gg d_{\mathrm{model}}$ is the inner size. These are used in the standard fashion by the network. The MHA parameters consist of 4 projection matrices ($\mathbf{W_q}$, $\mathbf{W_k}$, $\mathbf{W_v}$ and $\mathbf{W_o}$) of size $\mathbb{R}^{d_{\mathrm{model}} \times d_{\mathrm{model}}}$ (query, key, value, out). These are used to project the hidden vector to and from the component attention parts. In implementations, this projection is made with the matrices in their folded tensor form $\mathbb{R}^{n_{\mathrm{heads}} \times \frac{d_{\mathrm{model}}}{n_{\mathrm{heads}}} \times d_{\mathrm{model}}}$ where $n_{\mathrm{heads}}$ is the number of attention heads.
%where \textit{head} is refers to one of $n_{heads}$ projections.

In standard fine-tuning, starting from $\theta$, we optimize the loss $\cal{L}$ (for instance, cross-entropy for classification):
\[ \arg\min_{\theta'} \cal{L}(\theta') \]
Score-based pruning methods~\cite{scorebasedpruning} modify the model by introducing
score parameters $S$ for each parameter $i$ and replace the original
parameter matrices with a masked version $\mathbf{W}' = \mathbf{W} \odot M(\mathbf{S})$. For instance,
in the simplest version of magnitude pruning, the mask would just
zero-out parameters with low absolute values.

Movement pruning~\cite{sanh2020movement} is a score-based pruning approach that encourages the model to optimize these score
parameters. Specifically, we focus on the soft-movement variant of movement pruning that  sets $M(\mathbf{S}) = \mathbf{1}(\mathbf{S} > \tau)$ for a
threshold parameter $\tau$, and optimizes a regularized objective,
\[ \arg\min_{\theta', S} \cal{L}(\theta') + \lambda \lVert\sigma(S)\rVert \]
where $\lambda$ is a hyper-parameter, $\lVert A \rVert = \sum_{i,j} A_{i,j}$ and $\sigma$ is the sigmoid function.

This pruning objective encourages the model to fine-tune the parameters while lowering the scores of unimportant parameters and thus 
encouraging more sparsity. In order to train through the threshold, a
straight-through estimator~\citep{Bengio2013EstimatingOP} is used.
%This yields gradient updates of
%the form: \victor{Not sure we need this gradient form. Plus it requires to introduce more notation with the activation in particular}

%\[ \frac{\partial {\cal L}}{\partial W_{i,j}} = \frac{\partial {\cal L}}{\partial a_i} M_{i,j} x_j \]

Movement pruning, combined with distillation, has shown to be a very
effective method to reduce the number of parameters in an
existing model yielding 94\% pruning  in our tests for a F1 of 87.5 on SQuAD v1.1 (BERT-base is 88.5). This results in significantly smaller models than distillation alone.
However, even with this sparsity level, the model is not substantially faster when run on most standard hardware that cannot significantly take advantage of this style of sparse matrix-vector product.

\section{Model: Block Movement Pruning}

In this work, we extend movement pruning to work on blocks of local parameters. Specifically, each 
matrix in the transformer is partitioned into fixed-sized blocks. This setting goes beyond the arbitrary pruning of unstructured methods, with the goal of encouraging  the data locality closer to what would be needed for efficiency.\footnote{Linear algebra libraries perform matrix multiplication using large blocks, typically 128*64. At a micro level those machines are typically 32 ways SIMD, and memory is loaded by large contiguous chunks to maximize bandwidth.  Unstructured sparsity is hard to implement with dense algebra performance on GPUs. Data locality is important on CPU too, but in a more limited way.}

Our approach is extremely simple. 
For each parameter matrix $\mathbf{W} \in \mathbb{R}^{M \times N}$, we assume a
fixed-sized block structure $(M', N')$. Each of these blocks acts as
an individual group in the regularization with a shared score
parameter derived from the corresponding score matrix
$\mathbf{S} \in \mathbb{R}^{M/M' \times N / N'}$. Computing the
masked weight is done by expanding the thresholded values, i.e.
\[W'_{i,j} =  W_{i,j} * M(\mathbf{S})_{\left \lceil{i / M'}\right \rceil , \left \lceil{j / N'}\right \rceil } \]
As in past work, this model is trained with distillation to match the performance of a teacher model. 

Unlike other distillation approaches that require fully specifying the new model structure, our method only requires the size and shapes of the blocks, i.e. the set of
$(M',N')$ for each parameter matrix in the model. If blocks are too large, then they are difficult to prune, but if they are too small they do not support efficient inference.

To reduce the search
space, we will limit ourselves to test ${(M', N')^{\text{att}}}$ and
${(M', N')^{\text{ff}}}$: the same block size will be used for all layers for
attention weights $\mathbf{W_q}$, $\mathbf{W_k}$,
  $\mathbf{W_v}$ and $\mathbf{W_o}$ on one hand, and for the feed-forward weights $\mathbf{W_1}$ and
  $\mathbf{W_2}$ on the other hand. We split the movement pruning regularization term into:
\[ \lambda_{\mathrm{att}} \lVert\sigma(S_{\mathrm{att}})\rVert  + \lambda_{\mathrm{ffn}} \lVert\sigma(S_{\mathrm{ffn}})\rVert \]
This allows us to take into account the difference in terms of gradient received by the score parameters.

 To reduce further the search
space, we will test on two kinds of blocks:

\begin{itemize}
\item $(32, 32)$ : square blocks (\textit{Block})
\item $(1, d_{\mathrm{model}})$  and $(d_{\mathrm{model}},1)$ : dimension pruning on paired FFN rows and columns (\textit{Dim})
\end{itemize}

These block sizes allow for efficient models: blocks of size at least $(16,16)$ are efficient to compute with appropriate GPU kernels, whereas full rows, columns or heads can be entirely removed from the matrix: the remaining matrix is then dense.

We also include two additional baseline block types used to verify the approach:

\begin{itemize}
\item $(2^n, 2^n), n \in [2,5]$ : smaller power of two square block sizes to study the impact of size on performance (\textit{Block})
\item $(\frac{d_{\mathrm{model}}}{n_{\mathrm{heads}}}, d_{\mathrm{model}})$ : for attention heads (\textit{Heads})
\end{itemize}

The first considers small blocks, and the second considers very large functional blocks. 

\section{Experimental Setup}

We conduct experiments on five  (English) tasks commonly used to evaluate pre-trained language models: question answering \citep[SQuAD v1.1][]{rajpurkar2016squad} and \citep[SQuAD v2][]{rajpurkar2018know}, natural language inference \citep[MNLI][]{williams2017broad}, sentence similarity \citep[QQP][]{chen2018quora}, sentiment classification \citep[SST-2][]{socher2013recursive} and abstractive summarization~\citep[CNN/DailyMail][]{hermann2015teaching}. These datasets respectively contain 87k, 130k, 392k, 363k, 67k and 287k training examples, and are downloaded from the Hugging Face datasets hub. SQuAD is formulated as a span-extraction task, MNLI and QQP are sentence pairs classification tasks, SST-2 is a sentence classification task and CNN/DailyMail (“CNN”) is formulated as a conditional generation task. We report the performance on the development set as measured by the accuracy for MNLI and SST-2, F1 for QQP, the exact match (EM) and F1 for SQuAD and ROUGE for CNN/DailyMail.

We experiment with task-specific pruning of transformer language models. We use BERT \citep{devlin2018bert} (an encoder-only Transformer language model with 110M parameters, among which 85M are part of the linear layers present in the Transformer layers) for sentence classification and question answering (340M and 227M respectively for BERT-large), and BART \citep{lewis2019bart} (an encoder-decoder language model with 139M parameters, among which 99M are part of the linear layers present in the Transformer layers) for summarization (406M and 353M for BART-large). 

We compare against several baselines. Movement pruning is a fully unstructured approach and gives an upper bound on the sparsity trade-offs we hope to achieve, even if it provides little speed benefit. We also compare our results against state-of-the-art approaches developed for fast inference of transformer-based language models. DistilBERT \citep{sanh2019distilbert} is obtained by distilling through pre-training a pre-trained BERT into a smaller model. TinyBERT~\citep{jiao2019tinybert} distills a fine-tuned model while using data augmentation. MobileBERT~\citep{sun2020mobilebert} is the result of a large architecture search. dBART~\citep{shleifer2020pre} is obtained by arbitrarily copying equally spaced layers of a large model to a smaller one. To measure inference speed on GPU, we use a 24GB 3090 RTX and an Intel i7 CPU, using a large batch size (128) for evaluation and using PyTorch CUDA timing primitives.  We measure the speed of other models in this same setup. Results may be different from original papers, as latency and throughput characteristics are different for each platform. We also provide the number of parameters in the linear layers of the Transformer layers for each of our models and for the reference ones: as the linear layers represent most of the FLOPS, this is a good proxy for the computation required and to some extent for the compute time, when the model characteristics are equivalent.

\subsubsection*{Resources and Reproducibility}

We are using a minimal set of hyperparameters. The ratio of $\mathbf{\lambda_{att}}$ and $\mathbf{\lambda_{ffn}}$ is fixed by the relative sizes. We performed a few experiments with different values fixed manually for these parameters, but their influence is minor.

The main hyperparameter is the number of training epochs. For SQuAD v1.1, we are using 20 epochs instead of typically 2 for BERT models. This means a fine-tuning is taking about 12h with our method instead of 45mn with a standard fine-tuning setup. This number has to be large enough to let pruning happen slowly enough for a given task. A warming up phase and a post-pruning cool-down phase are helpful, but their exact length has not a large impact on final performance.
We believe the training time is less important than the inference time for energy consideration, as inference is performed repeatedly. Our method is optimizing inference by a large factor: the training energy is potentially recouped by a large margin with inference savings.

Finally, the checkpoints created during the experiments are available on an AWS S3 bucket, with their metadata and  training parameters, totaling 3TB of data, to facilitate reproduction of our results and to make it possible to study further the behavior of those models. Code for experiments, analysis, and tools to prepare the present paper are available on GitHub (see Appendix \ref{appendix:code}). 

\subsubsection*{Pruning Methods}

\begin{table}[t]
\centering
\begin{tabular}{ @{}lccc@{} }
\toprule
Method & MHA & FFN & Teacher\\
\midrule
Block & Block & Block & Base\\
Hybrid & Block & Dim & Base\\
Hybrid NT & Heads & Dim & None \\
Struct & Heads & Dim & Base\\
Hybrid Filled & Heads & Dim & Base\\
Hybrid Filled LT & Heads & Dim & Large\\
\bottomrule
\end{tabular}
\caption{Summary of pruning methods. Dim blocks correspond to  row and column blocks for the FFN.}
\label{table:pruning_methods}
\end{table}

The pruning approaches are shown in Table~\ref{table:pruning_methods}.

\textbf{Block} pruning use square block sizes throughout all the linear layers, as an extension of the original movement pruning for which the block size is 1. 

\textbf{Hybrid} pruning jointly removes hidden dimensions in feed-forward layers $\mathbf{W_1}$ and $\mathbf{W_2}$, using movement pruning to create the dimension mask. This corresponds to full rows or columns in the parameter matrices. The pruned $\mathbf{W'_1}$ and $\mathbf{W'_2}$ can then be "compacted" to become fully dense:  we perform dense operations on cropped matrices. For the attention layers, pruning only some rows or columns in $\mathbf{W_q}$, $\mathbf{W_k}$, $\mathbf{W_v}$ and $\mathbf{W_o}$ can not be practically exploited. This is because the structure of the computation makes the additional cost of resizing the tensor inefficient.  We, therefore, use square block pruning on the attention layer, with a block size of $(32, 32)$ which showed the best tradeoff between performance and accuracy.

\textbf{Struct} pruning uses the same methods for FFN layers but aims to remove model attention heads directly. To do so, we choose a block size on attention that equals the head size while still using the same soft movement pruning strategy. For this approach, we use a $\lambda_{att}$ equals to $1/32$, as there are 32 times more parameters than in an attention block than in a feed-forward dimension.

When Block Pruning does not fully remove a component such as an attention head, as shown in Figure~\ref{fig:pruning_patterns}, we cannot speed up the model. But we can reclaim some of the performance at no speed cost and at marginal cost on sparsity by making use of those zero weights.

\textbf{Hybrid Filled} pruning allows the model to reinitialize these reclaimed weights uniformly at random and continue fine-tuning the smaller model for a few steps. We also explore "rewinding"~\citep{LotteryTicketHypothesis} by identifying weights that should not be pruned (because they are part of a non-empty attention head) and re-fine-pruning the pre-trained model: the first run marks the attention heads that were not pruned, and the second uses this information to create a positive mask of weights that are protected from pruning. We did not find a significant difference between the two methods. The results presented here do not use rewinding.

\section{Experiments}
\label{section:experiments}

\newcommand{\fullpagefigure}[3]{{
\begin{figure*}[t]
\includegraphics[width=\textwidth]{images/#2}
\caption{#3}
\label{fig:#1}
\end{figure*}
}}

\newcommand{\singlecolumnfigure}[3]{{
\begin{figure}[t]
\centering
\includegraphics[width=\columnwidth]{images/#2}
\caption{#3}
\label{fig:#1}
\end{figure}
}}

\newcommand{\layerpatterntext}{{
\small
\begin{minipage}[b]{0.2\columnwidth}
Attention
\end{minipage}
\begin{minipage}[b]{0.8\columnwidth}
\hspace*{1mm}Feed-forward
\end{minipage}
}}

\newcommand{\layerpattern}[4]{{
\begin{minipage}[b]{1.0\columnwidth}
\small
#4
\end{minipage}
\begin{minipage}[b]{0.2\columnwidth}
\includegraphics[width=\columnwidth]{images/attention_#2.pdf}
\end{minipage}
\begin{minipage}[b]{0.8\columnwidth}
\includegraphics[width=\columnwidth]{images/ff_#3.pdf}
\end{minipage}
\newline
}}

\newcommand{\pruningpatterns}{{
\begin{figure}[t]
\layerpattern{block_pattern}{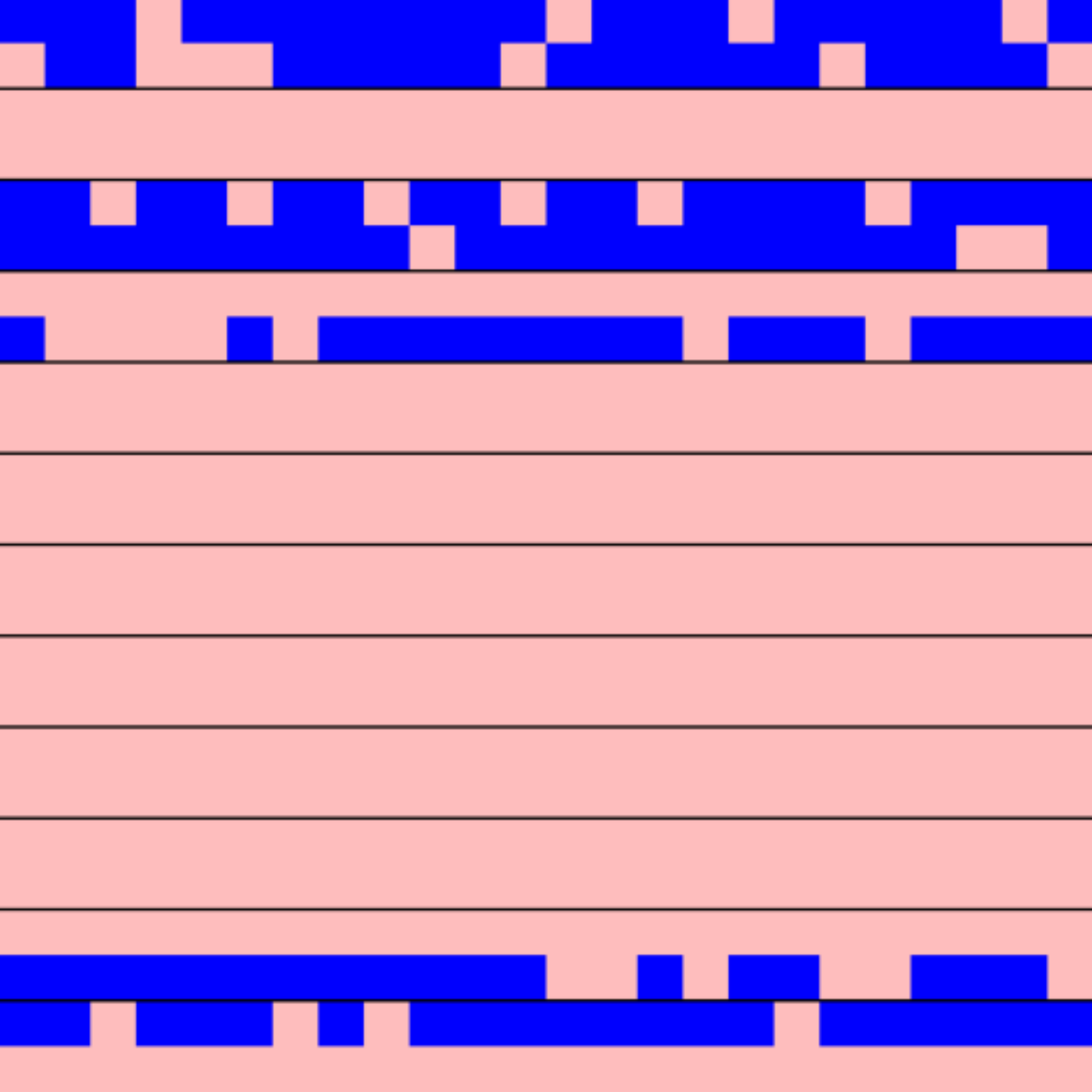}{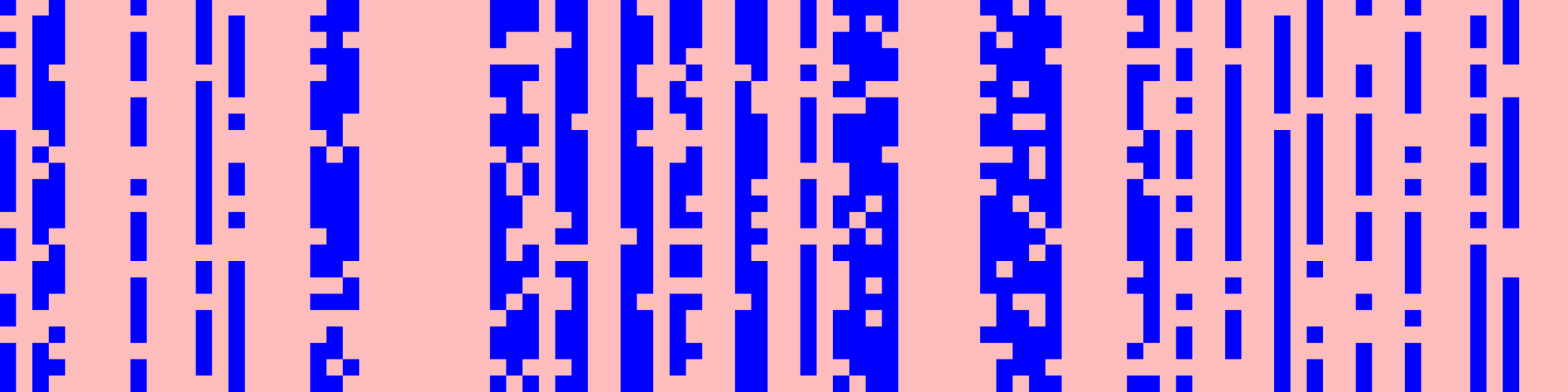}{Block}
\layerpattern{hybrid_pattern}{block}{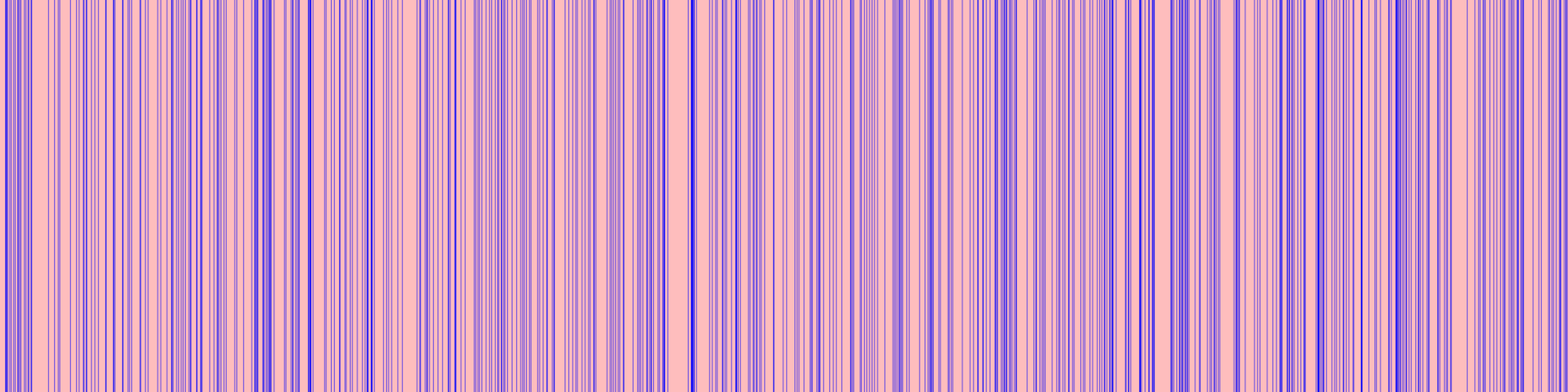}{Hybrid}
\layerpattern{hybrid_filled_pattern}{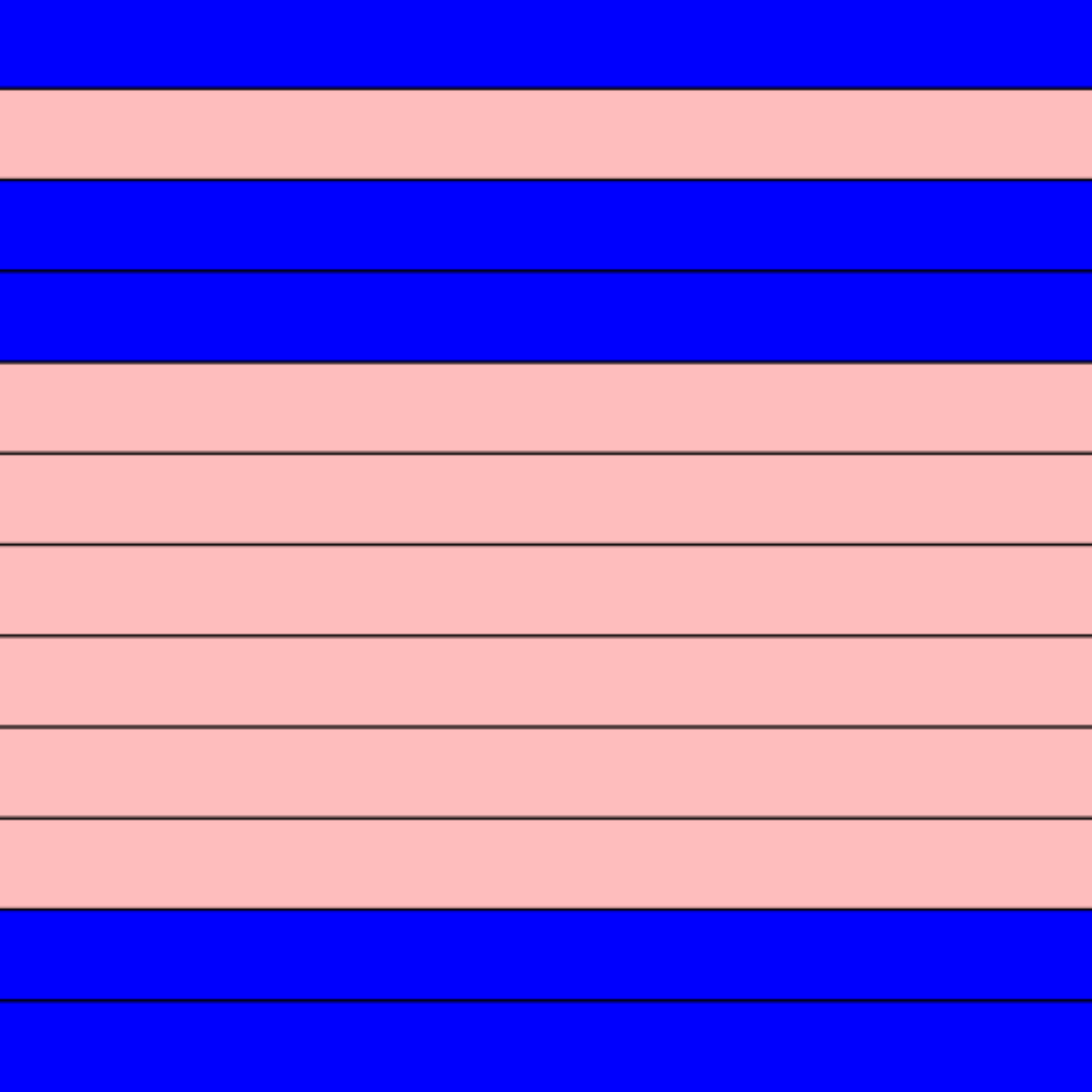}{col}{Hybrid Filled}
\layerpatterntext
\caption{Pruning patterns on SQuAD v1.1: blue is preserved, pink is pruned. Attention heads are delimited for clarity. }
\label{fig:pruning_patterns}
\end{figure}
}}

%We use SQuAD v1.1 as the main task to test different block size choices for attention and feed forward layers, and validate our results on other tasks. 
\paragraph{Main Results}

We begin by observing the high-level impact of the different pruning methods. 
Figure~\ref{fig:pruning_patterns} shows the effect on attention and feed-forward layers for the different block pruning methods. We find that all the different block sizes learn to prune out entire dimensions in the FFN layers. Interestingly we find that the block methods can also learn to remove entire heads from the MHA. This pruning pattern makes it possible to remove entire heads from the model during inference. For this reason, we focus on the Hybrid approach as our main method, which can both eliminate feed-forward dimensions while using blocks to remove attention heads gradually.

\pruningpatterns

\begin{figure*}
    \centering
    \includegraphics[width=0.98\columnwidth]{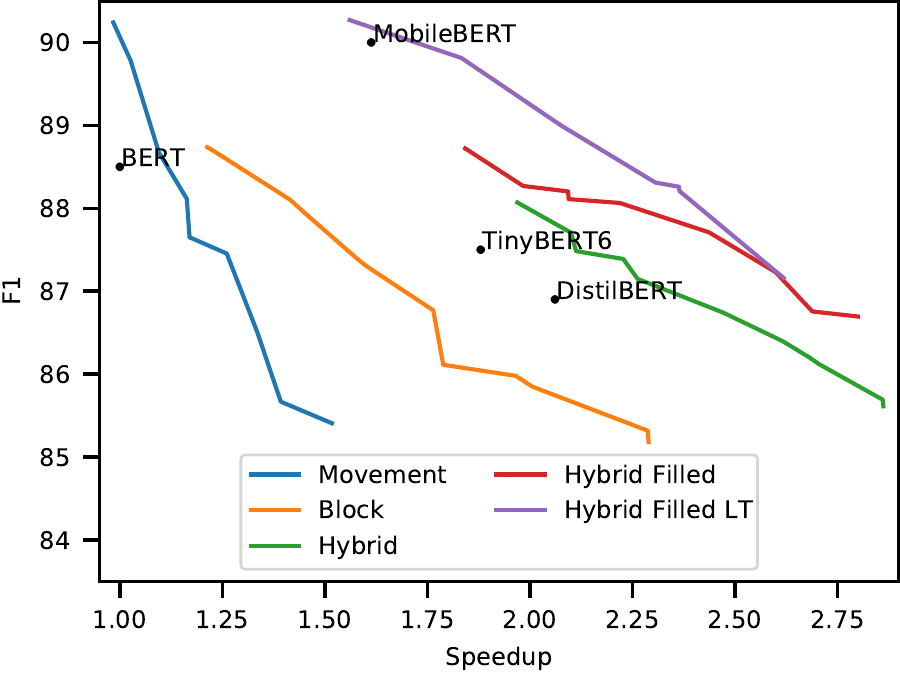}\hspace*{0.5cm}\includegraphics[width=0.98\columnwidth]{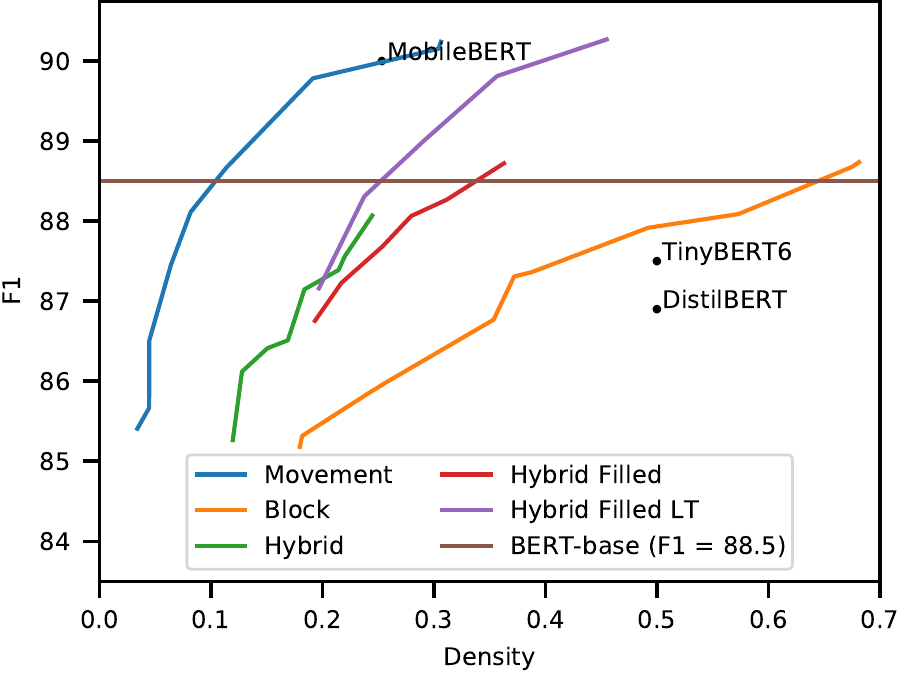}
    \caption{Comparison on SQuAD v1.1 of model F1 against speed and density (BERT-base reference). For each pruning method the pruned model is BERT-base, but different regularization values give different final sparsity levels. This translates into a tradeoff curve between accuracy and speedup specific to the method. Distilled networks (Mobile|Tiny|Distil)BERT are given as references. The higher the curve, the most accurate the model is for a given speed. }
    \label{fig:squadv1_summary}
\end{figure*}

%\singlecolumnfigure{squadv1_summary}{squadv1/paper_summary_speedup}{}

\begin{table}[t]
\small
\begin{tabular}{ @{}llllll@{} }
\toprule
Model & Size & Compr. & Speed & F1 & EM\\
\midrule
BERT & 85M & 1.00 & 1.00 & 88.5 & 80.8 \\
TinyBERT & 42M & 2.00 & 1.88 & 87.5 & 79.7\\
DistilBERT & 42M & 2.00 & 2.06 & 86.9 & 79.1\\
Hybrid Filled & 30M & 2.76 & 1.84 & 88.7 & 81.7\\
Hybrid Filled & 21M & 3.94 & 2.25 & 87.7 & 80.2\\
Hybrid & 20M & 4.09 & 1.97 & 88.07 & 80.6\\
Hybrid Filled & 16M & 5.17 & 2.69 & 86.8 & 78.8\\
Hybrid & 15M & 5.44 & 2.26 & 87.15 & 79.4\\
Hybrid & 14M & 5.91 & 2.24 & 86.51 & 78.7\\
Struct & 14M & 6.25 & 3.36 & 85.9 & 78.0\\
Hybrid & 12M & 6.63 & 2.61 & 86.41 & 78.3\\
Hybrid & 10M & 7.82 & 2.70 & 86.12 & 77.8\\
\midrule
MobileBERT & 21M & 4.00 & 1.61 & 90.0 & 82.9\\
Hybrid Filled LT & 38M & 2.20 & 1.56 & 90.3 & 83.7\\
Hybrid Filled LT & 20M & 4.21 & 2.31 & 88.3 & 81.2\\
Hybrid Filled LT & 16M & 5.08 & 2.62 & 87.2 & 79.5\\
\bottomrule
\end{tabular}
\caption{SQuAD v1.1 pruned models with base/large teacher: Encoder Linear Parameters Count, Compression rate and Speed gain relative to BERT-base, F1 and Exact Match. LT stands for Large Teacher.}
\label{table:squadv1}
\end{table}

Results on SQuAD are shown in Figure~\ref{fig:squadv1_summary}, which compares our approach for speed and density
to baseline BERT-Base tuned models such as TinyBERT-6 and DistilBERT (MobileBERT is discussed below). 
The main result is that the Hybrid Pruning model is as fast as the baseline and approaches the same accuracy while at the same time producing significantly smaller models in terms of density. 
Moving to the Hybrid Filled model leads to a further gain in speed at a small cost in model density. 
 For instance, for the same F1 performance of 87.5, Hybrid Filled models display a 2.5x speedup against 1.88 for TinyBERT. TinyBERT and DistilBERT have 50\% of BERT's encoder parameters, whereas Hybrid Filled models have 25\% BERT parameters for the same level of accuracy. 

The figures also include two intrinsic baselines: our reimplementation of Movement pruning and pure Block pruning. We find that our implementation of Movement pruning is highly effective at producing sparse models (even leading to a small increase in accuracy) but does not produce significant speedups. Square Block pruning does better, but not as well as hybrid blocks.

%For Block Pruning,  we observe speed gains over the movement pruning baseline, even though DistilBERT and TinyBERT are still faster for a comparable accuracy. Hybrid Pruning results in a much higher speedup for the same accuracy and a higher sparsity, exceeding TinyBERT performance on SQuAD v1, both in terms of F1 and speed.

%\singlecolumnfigure{squadv1_summary_fill_rate}{squadv1/paper_summary_fill_rate}{Comparison on SQuAD v1.1 of model F1 against density (BERT-base reference).}

%Figure~\ref{fig:squadv1_summary_fill_rate} shows that the size difference is even larger: 

 Table~\ref{table:squadv1} gives a full comparison of models with different compression rates. As linear layers represent a very large part of the flops of a transformer model, this compression rate is actually a good measure of the maximum achievable speedup. This number is much higher than the actually measured speedup. This indicates that our setup for measuring speed may underestimate the actual speedup one could obtain with those pruned models with specialized implementations. Hybrid Filled reaches a 2.25x speedup under minimal loss in accuracy. Struct pruning targeting MHA blocks directly can be even faster but leads to a stronger degradation in accuracy.

\begin{table}[t]
\small
\begin{tabular}{ @{}lllll@{} }
\toprule
Model & SQuAD v1.1 & MNLI & QQP & SST-2\\
 & F1/EM & Acc (m/mm) & F1 & Acc\\
\midrule
TinyBERT & 87.5/79.7 & \textbf{84.6}/83.2 & \textbf{88.0} & \textbf{93.0}\\
Hybrid & \textbf{88.1}/\textbf{80.6} & 83.2/\textbf{83.6} & 87.9 & 91.2\\
%Model Density & 24\% & 27\% & 36\% & 38\% \\
\bottomrule
\end{tabular}
\caption{Hybrid pruning/TinyBERT cross-task performance comparison. Speed is \textit{at least} TinyBERT speed (1.88x BERT-base) for all networks and significantly sparser.}
\label{table:tasks_hybrid_tinybert}
\end{table}

% SRUSH: I think this information is already clear in the graph
% \begin{table}[t]
% \small
% \begin{tabular}{ @{}lrrr@{} }
% \toprule
% Speedup & 2.0x & 2.2x & 2.6x\\
% \midrule
% Hybrid & 88.1 & 87.4 & 85.3\\
% Hybrid Filled & 88.3 & 88.1 & 87.2\\
% Gain & 0.2 & 0.7 & 2.0\\
% \bottomrule
% \end{tabular}
% \caption{Hybrid Filled gain over non filled network on SQuAD v1. Models speeds are chosen equal.}
% \label{table:filled_gain}
% \end{table}

Table~\ref{table:tasks_hybrid_tinybert} shows the comparison between TinyBERT and a Hybrid pruned model of the same speed on several others tasks. Hybrid Pruning performs better on SQuAD v1.1, and approaches TinyBERT performance on other tasks. 
%It should be noted that the results here are not using Hybrid Filled which would improve those results.\fixedcomment{describe those results}

% as seen in Table \ref{table:filled_gain}: as sparsity increases, more attention heads are pruned, but the remaining heads are themselves more sparse, so re-filling them gives back more performance in comparison.

\paragraph{Comparison with MobileBERT}
All methods can be improved further using a larger teacher model. For these experiments, we compare with 
MobileBERT, which uses a BERT-large teacher and reaches an F1 of 90.0 on SQuAD v1.1 on its fastest version. It should be noted that MobileBERT makes use of additional optimizations not present in the original BERT-large we are using: LayerNorms are replaced by purely linear NoNorms, and GeLUs are replaced by ReLUs.
For these experiments, we use a BERT-large teacher to perform meaningful comparisons, using our best method Hybrid Filled.

Figure~\ref{fig:squadv1_summary} shows that we have comparable results on SQuAD v1.1, with a simpler optimization approach: we get a slightly better model (F1=90.3) for the same speedup of 1.6x, and we get a speedup of 2.2x at BERT-base accuracy (F1=88.5).
We observe that using a large teacher is beneficial even at high levels of pruning: up to 80\% of sparsity, the resulting student model has better accuracy for the same number of parameters when using a BERT-large teacher instead of a base one. This trend reverses after this point: a larger teacher is detrimental to accuracy when the student is very heavily pruned.

%Finally, a MobileBERT \cite{sun2020mobilebert} claim is that latency can be improved by a 3.1x factor on mobile devices, thanks to the use of NoNorm instead of LayerNorm, and the use of ReLUs instead of GeLUs. As an additional benefit, we designed a method to replace gradually existing BERT LayerNorms into NoNorms, and to replace GeLUs by ReLUs in a similar way. The architectures are then strictly feature equivalent.

%For both experiments, Hybrid Filled Pruning would probably improves the performance of the models significantly.

\paragraph{Encoder-Decoder}

%There has been a significant amount of research on structured pruning of encoder-only Transformer language model ~\citep{StructuredPruningQA, kim2020fastformers}.

Finally, we apply these methods to two encoder-decoder architectures,  BART-base and BART-large for the task of summarization. For these architectures, the decoder parameters are responsible for a majority of the computational costs, so these are our main focus. \citet{voita2019analyzing} observed that for machine translation models, encoder heads were much easier to prune than decoder ones. We found similar results, e.g. for identical $\mathbf{\lambda_{att}}$ and $\mathbf{\lambda_{ffn}}$, the encoder was systematically more pruned than the decoder, for both MHA and FFN sub-layers.
 In order to increase speedup gain, we applied twice as much weight on the decoder compression, which resulted in even pruning ratios among the encoder and decoder.

Table~\ref{table:bartcnn} shows the main results. We see that Hybrid pruning leads to large decoder compression ratios (3.4 on BART-base and 3.5 BART-large) with only a small drop in ROUGE score. Speedups reach 1.4 times of the original speed. (Given the large decoder compression rates, we would expect larger speedups to be possible with further engineering of the inference.)

There is less comparable work for pre-trained encoder-decoders. 
We compare our approach with a distillation-based approach dBART~\citep{shleifer2020pre}. 
This approach yields a similar speedup gain with a smaller drop in performance but less sparsity.
For models of comparable sizes (158M for our Hybrid NT vs 176M for dBART-6-6), we observe a drop of 0.7 in R2 and 0.4 in RL against  0.9 in R2 and 1.3 in RL for dBART-6-6. As with encoder-only models, the two approaches could likely be combined to yield even faster, smaller models.\footnote{Distillation methods for text generation require generating pseudo-labels,  a different process which is significantly slower than BERT distillation.}

% Interesting experiment  would have been : pruning only decoder 

\begin{table}[t]
% \small
\resizebox{\linewidth}{!}{
    \begin{tabular}{ @{}llllllll@{} }
    \toprule
    Model & Size & DCp & Speed & R1 & R2 & RL \\
    \midrule
    BART-large   & 353M & 1.0 & 1.00 & 44.8 & 21.7 & 41.8 \\  % 100%
    %Hybrid NT & 268M & 1.3 & -    & 44.9 & 21.6 & 42.1 \\  
    Hybrid NT & 158M & 2.0 & -    & 44.3 & 21.0 & 41.4 \\  %  44.73%
    Hybrid NT & 108M & 2.8 & 1.38 & 43.5 & 20.3 & 40.6 \\  %  30.49%
    Hybrid NT &  82M & 3.5 & 1.39 & 42.7 & 19.6 & 39.9 \\  %  23.27%
    \midrule
    BART-large$^{\dagger}$  &  353M & 1.0 & 1.00 & - & 21.1 & 40.9 \\ 
    dBART-12-6$^{\dagger}$  &  252M & 2.0 & 1.44 & - & 21.2 & 41.0 \\ 
    dBART-6-6$^{\dagger}$    &  176M & 2.0 & 1.46 & - & 20.2 & 39.6 \\ 
    dBART-12-3$^{\dagger}$   &  201M & 4.0 & 1.66 & - & 20.6 & 40.3 \\ 
    \midrule
    \midrule
    BART-base         & 99M & 1.0 & 1.00 & 43.4 & 20.4 & 40.4 \\  % 100%
    Hybrid NT  &  35M & 2.6 & 1.19 & 42.2 & 19.4 & 39.2 \\  %  35%
    Hybrid NT  &  23M & 3.4 & 1.35 & 41.4 & 18.7 & 38.4 \\  %  23.18%
    \bottomrule
    \end{tabular}
}
\caption{BART pruned models fine-tuned on CNN: encoder-decoder linear parameters count, Decoder compression rate (DCp) and Speed gain (one forward prediction) relative to BART-large/base, dev ROUGE scores. $\dagger$ denotes test ROUGE scores taken from \citet{shleifer2020pre}. NT stands for No Teacher.}
\label{table:bartcnn}
\end{table}

\section{Analysis}

\paragraph{Large Model Pruning}

\begin{table}[t]
\small
\begin{tabular}{ @{}lllllll@{}}
\toprule
Model & Size & Compr. & Speed & F1 & EM\\
\midrule
SQuAD v1.1\\
BERT-large & 227M & 1.00 & 1.0 & 93.2 & 86.9\\
BERT-base & 85M & 2.66 & 3.1 & 88.5 & 80.8\\
Hybrid & 54M & 4.16 & 2.9 & 91.0 & 84.6\\
Hybrid & 41M & 5.59 & 3.2 & 90.2 & 83.7\\
\midrule
SQuAD v2\\
BERT-large & 227M & 1.00 & 1.0 & 85.8 & 82.8\\
StructuredQA & 57M & 4.00 & -- & 81.5 & -- \\
Hybrid  & 38M & 5.88 & -- & 82.6 & 79.7\\
\bottomrule
\end{tabular}
\caption{BERT-large SQuAD pruned models. Reference speed is BERT-large.}
\label{table:squadv1_large}
\end{table}

To test that this approach scales to large models, we apply Hybrid pruning on BERT-large on SQuAD v1.1. We observe similar results: a 18\% dense BERT-large has a F1 of 90.2, with a speedup of 3.2x compared to BERT-large with a F1 of 93.2. This pruned model is actually faster than a BERT-base model (Table~\ref{table:squadv1_large}). 
We can compare Hybrid Pruning of SQuAD v2 BERT-large models with the results of the structured pruning method described in \citet{StructuredPruningQA}. For a 17\% dense model, we obtain a F1 of 82.6, whereas structured pruning gets a 25\% dense model with a F1 of 81.5.

This result is in line with~\citet{trainlargethencompress}: the larger the model, the more pruning is effective. When pruning a larger model, the final model is actually better than a smaller one with the same absolute number of parameters.

\paragraph{Block Size Influence}

\singlecolumnfigure{squadv1_blocks}{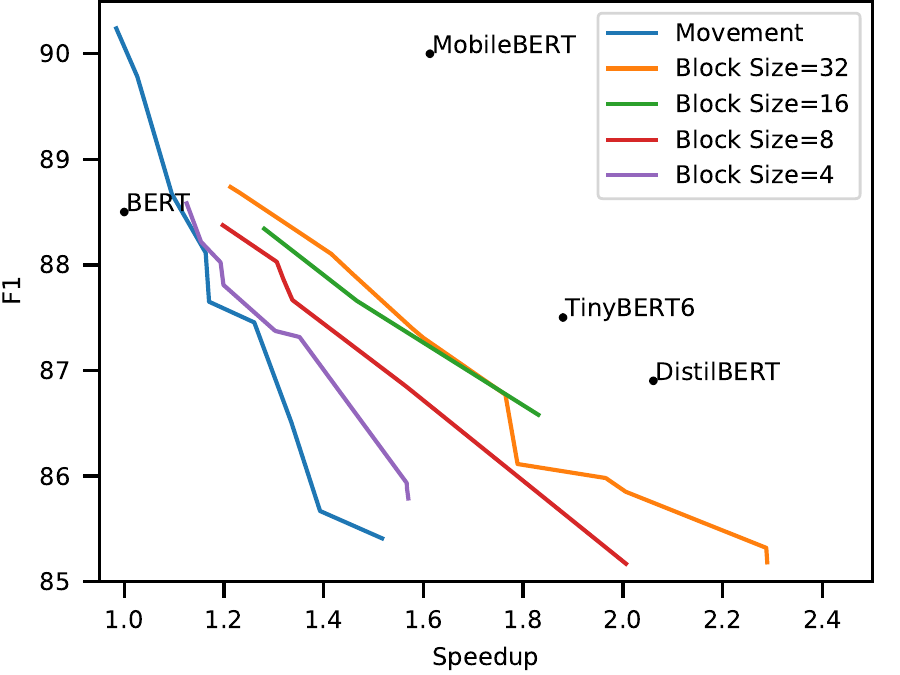}{SQuAD v1.1 with Block Pruning: Influence of block size on F1.}

Figure~\ref{fig:squadv1_blocks} shows the impact of different block sizes on Block pruning: pruning is done on attention layers and FFNs with the same square block size, from $(4,4)$ to $(32,32)$, with a BERT-base teacher.
We can see that we reach the BERT-base original F1 for all block sizes from 4 to 32, but with a speedup that increases with the block size. The maximum reachable speedup without F1 drop is 1.3 for a block size of 32. But when some drop of F1 is allowed, the speedup increases quickly with the block size and plateau when reaching 16. We then reach a speedup of 1.75 for an F1 drop of 2\% and a block size of 32.

We also note that, with the original Movement Pruning method, we see some speedup due to full dimension pruning.  This likely comes from our improved set of hyper-parameters (compared to the original paper), allowing us to remove some empty rows and columns in the FFN layers. 
However we see that using blocks leads to a significant speed improvement compared to Movement Pruning.

%When observing the "Block" section in Figure \ref{fig:pruning_patterns} we see why the models are faster: the block pattern is far from random. The larger the blocks the most probable full rows or columns will disappear, especially for attention layers.

\paragraph{Quantization}
Quantization is often of critical importance for practical applications. We, therefore, wanted to check that our networks could be subjected to quantization without significant loss of accuracy, especially when considering the issues that could arise with the high level of sparsity of some FFNs.
Table~\ref{table:bertquant} shows the results of full 8-bit quantization tests on our models. These indicate that the method is compatible with quantization, and the models using quantization on top of our pruning method achieve very high gains in terms of size (as well as speed). 
%This is especially true when using the MobileBERT optimizations \todo{cite APPENDIX}, making quantization even easier and effective.

\begin{table}[t]
\centering
\small
\begin{tabular}{ @{}llllllll@{} }
\toprule
Model & Compress & EM & F1 \\
\midrule 
BERT-base        & 1.0  & 80.8 & 88.5 \\ 
Hybrid           & 3.3 & 80.2 &  87.8 \\ 
+ quantization   & 13.3 & 77.8 & 86.3 \\  
\bottomrule
\end{tabular}
\caption{Results of pruned BERT-base on SQuAD v1.1 dev Exact Match and F1 score with 8-bit quantization.}
\label{table:bertquant}
\end{table}
\paragraph{Impact of Distillation}

We report experimental results with the addition of a teacher distillation step as 
previous work showed this boosts movement pruning at little cost.
In this section, we conduct an ablation study to evaluate the impact of distillation using a BERT-base teacher.

As shown in Table~\ref{table:ablation_study}, combining hybrid pruning with distillation always performs better than pruning alone, but that it is not critical for the approach to work.
The distillation effect is larger for smaller datasets such as SST-2, which are prone to over-fitting. We believe that the regularization brought by pruning and distillation counters over-fitting caused by the additional number of steps needed for pruning.  

%The difference in performance shows that our hybrid method is not dependant of distillation, but that the latter helps improving performances. 

\begin{table}[t]
\small
\centering
\begin{tabular}{ @{}lccc@{} }
\toprule
Dataset & Size & Hybrid  & Hybrid NT \\
\midrule
QQP &24M & 87.61 & 87.17  \\
& 21M & 87.14 & 87.00  \\
& 10M & 86.82 & 86.27  \\
\midrule
SST-2 & 70M & 93.23 & 92.20 \\
& 42M & 91.97 & 90.71  \\
& 18M & 90.60 & 89.79  \\
\bottomrule
\end{tabular}
\caption{Distillation ablation study of BERT-base on QQP and SST-2 dev of a BERT-base teacher. F1 and accuracy scores reported for QQP and SST-2 respectively. NT stands for No Teacher.}
\label{table:ablation_study}
\end{table}

% We introduced block movement pruning because it was a sensible way to achieve a good performance on parallel hardware, whereas unstructured pruning is difficult to accelerate.
% We then observed an emerging effect: using blocks increases the probability that full structures disappear entirely.
%It terms of training, the transition is smoother if a single block disappears than if a entire heads does, as is usually done in pruning methods. But the end result is very similar: full heads are removed, and the remaining ones have a high density.

% The produced models are faster than the BERT-base distilled ones, when using the complete Hybrid Filled method. But their size is even much lower, while being still dense. The speed we measure is probably underestimated compared to what the best production setups would give. 
% MobileBERT is still smaller, but not faster because of its 24 layers, making it longer to compute on parallel hardware. 
% On mobile devices, MobileBERT claims a large speedup over BERT thanks to NoNorms and ReLUs. But as mentioned earlier, we can transform the original BERT model while fine-pruning: the final model should have performance very similar to MobileBERT on mobile devices, only the layer sizes are different.

\section{Conclusion}
We have shown that we can extract small pruned models that are at an equivalent or better than distilled networks. This approach can be done during fine-tuning and not pre-training. The method does not resort to techniques such as data augmentation or architecture search, and it works on a diverse set of tasks and base models. As better and larger models are published at an increasing pace, we can rely on a simple and robust method to accelerate them on specific tasks without sacrificing accuracy and  distribute these models easily while keeping most of the original model accuracy.

\section{Impact}
We expect the method presented here to contribute to the reduction of the compute resources and energy needed to perform natural language tasks, while preserving the original model performance. 
It will contribute additionally to alleviating privacy concerns: smaller models running on user devices instead of server-side allow more information to stay private. 
This is especially relevant when considering the large anticipated demand for such NLP applications in the near future.

\subsection{Acknowledgements}
The authors would like to thank the anonymous reviewers, the Hugging Face team for the support, Nvidia for providing us some hardware for evaluation, and finally the open-source community for the numerous tools which made this research possible.

\bibliographystyle{emnlp/acl_natbib}
\bibliography{main}
\clearpage
\appendix

\section{Reproducibility \& Hyper-Parameters}
\label{sec:appendix_hyperparameters}

\subsection*{Code}
\label{appendix:code}
The complete code to run the experiments, analyze the results and finally create the figures and tables in this paper is available on the Hugging Face nn\_pruning repository, at \uurl{https://github.com/huggingface/nn\_pruning}.

\subsection*{Hyperparameters}
The hyperparameters of the experiments are available as JSON files (one file per task) in the same repository: each entry contains all the information to fine-tune and prune the model, its evaluation results, and detailed statistics about its final sparsity.

For example, the SQuAD V1 checkpoints referenced in this paper are listed with the \href{https://github.com/huggingface/nn\_pruning/blob/main/analysis/article/files/results_squadv1.json}{hyperparameters and related information}.

\subsection*{Checkpoints}
 Some of the models we produced during this research can be used directly from the  \href{https://huggingface.co/madlag}{Hugging Face model hub}.
 
The other models and the checkpoints, including the intermediary ones that were saved during training, are \href{https://github.com/huggingface/nn_pruning/tree/main/analysis/article}{available on Amazon S3}.

\section{Additional Data}
\subsection*{Block Shape \& Head pruning}

We show here the effect of the pattern on the head number reduction: using block instead of row/column pruning leads to a much larger number of pruned heads while improving accuracy, here on the SST-2 task.

We are using Block Movement pruning for each model, with different block patterns, pruning only the attention layers. Compression measures the reduction of the number of non-zero parameters in attention linear layers, whereas head compression measures the reduction of the number of complete non-zero heads.

\begin{table}[!htbp]
\small
\begin{tabular}{ @{}lllll@{}}
\toprule
Pattern & Compr.  & Heads & Head Compr. &  Accur. \\
\midrule
BERT base  &    1x  &  144  &   1x  & 92.7 \\
\midrule
Rows/Cols & 8.6x & 86 &  1.7x  & 90.6 \\
Block 32  & 4.7x  &   54 &  2.7x &  91.1 \\
Block 64  & 3.5x  &   51  & 2.8x &  92.0 \\
\bottomrule
\end{tabular}
\caption{Head pruning method comparison on SST-2}
\label{table:head_pruning_comparison}
\end{table}

\subsection*{Pruning Methods Comparison}

We select speed as our main metric to compare with other techniques, as it is the major practical measure of inference efficiency. On this metric, we decided to compare our models to the best models available i.e. the distilled models (MobileBERT, TinyBERT), even though the method is different, as they are the strongest "speed/accuracy" baseline available.

In Table~\ref{table:distillation_pruning_comparison} we compare  \citet{wang2019structured} with TinyBERT \cite{jiao2019tinybert} and MobileBERT \cite{sun2020mobilebert}.

\begin{table}[!htbp]
\small
\begin{tabular}{ @{}llllll@{}}
\toprule
Method      &        Speed  & SST2  & MRPC &  STS-B &  QNLI \\
\midrule
Wang et al. &  <1.5x & 92.09 & 88.61 & 88.18 & 89.05 \\
TinyBERT           &   2x &  93.1 &  87.3 &  83.7  & 90.4 \\
MobileBERT        & 4x & 92.8 &  88.8 &  84.4 &  90.6 \\

\bottomrule
\end{tabular}
\caption{Distillation/Structured Pruning Comparison}
\label{table:distillation_pruning_comparison}
\end{table}

We compare as well to Hybrid pruning, with and without a teacher, with the unstructured methods from \citet{sanh2020movement} (the original Movement Pruning method we are using) and \citet{gordon2020compressing}, and with \citet{sajjad2020poor} (dropping full layers), in Table~\ref{table:pruning_method_comparison}.

\begin{table}[!htbp]
\small
\begin{tabular}{ @{}llllll@{}}
\toprule
Model   &     Spd & Cp  &      MNLI  & QQP &  SST-2 \\
& & &(m/mm) &  F1/Acc &  Acc  \\
\midrule
BERT base  &     1  &  1  &  84.5/85.1  & 88.1/91.1 & 92.7 \\
\midrule
Mvmt NT &     $\sim{1}$  & 10 &   80.7/81.1 & 87.1/90.5 &  -  \\
Hybrid NT   &  3.5  & 10 &   79.4/79.9 & 86.0/89.3 & 87.0 \\
Mvmt    &     $\sim{1}$  & 10 &   81.2/81.8 & 86.8/90.2 &  -  \\
Hybrid      &  3.5  & 10 &   80.4/81.1 & 86.4/89.8 & 89.7 \\
\midrule
Hybrid NT   &    3  & 4.5 &    81.7/81.8 &  87.0/90.3 & 89.8 \\
Hybrid      &    3  & 4.5 &    82.7/82.8 &  87.4/90.6 & 90.6 \\
Sajjad  & < 2 &   2 &   81.1/  -   &     - /90.4 & 90.3 \\
Gordon  &   -   & 2  &  82.6/- &   - /90.3 &  90.8 \\
Hybrid NT &   1.6  &  2  &  83.2/83.3 & 87.2/90.4 & 90.7 \\
Hybrid    &   1.6  &  2  &  83.7/84.1 & 88.3/91.3 & 92.0 \\

\bottomrule
\end{tabular}
\caption{Pruning Methods Comparison. Mvmt is Movement Pruning~\citet{sanh2020movement}, Sajjad is \citet{sajjad2020poor}, Gordon is \citet{gordon2020compressing}. NT is for No Teacher. Spd is speed multiplier, Cp is for parameters compression rate.}
\label{table:pruning_method_comparison}
\end{table}

\end{document}